\documentclass[conference]{IEEEtran}
\usepackage{epsfig}
\usepackage{graphicx}
\usepackage{amsmath}
\usepackage{amssymb}
\usepackage{subfig}
\usepackage{dblfloatfix} 
\usepackage{float}
\usepackage{color}
\usepackage[table]{xcolor}
\usepackage[pagebackref=true,breaklinks=true,letterpaper=true,colorlinks,bookmarks=false]{hyperref}

\begin{document}
\title{A fine-grained approach to scene text script identification}

\author{\IEEEauthorblockN{Llu\'is G\'omez and Dimosthenis Karatzas}
\IEEEauthorblockA{Computer Vision Center, Universitat Aut\`onoma de Barcelona\\
Email: \{lgomez,dimos\}@cvc.uab.es}}
\maketitle

\begin{abstract}
This paper focuses on the problem of script identification in unconstrained scenarios.
Script identification is an important prerequisite to recognition, and an indispensable condition for automatic text understanding systems designed for multi-language environments. Although widely studied for document images and handwritten documents, it remains an almost unexplored territory for scene text images. 

We detail a novel method for script identification in natural images that combines convolutional features and the Naive-Bayes Nearest Neighbor classifier. The proposed framework efficiently exploits the discriminative power of small stroke-parts, in a fine-grained classification framework.

In addition, we propose a new public benchmark dataset for the evaluation of joint text detection and script identification in natural scenes. Experiments done in this new dataset demonstrate that the proposed method yields state of the art results, while it generalizes well to different datasets and variable number of scripts. The evidence provided shows that multi-lingual scene text recognition in the wild is a viable proposition. Source code of the proposed method is made available online.

\end{abstract}


\IEEEpeerreviewmaketitle

\section{Introduction}
\label{sec:intro}  

Script and language identification are important steps in modern OCR systems designed for multi-language environments. Since text recognition algorithms are language-dependent, detecting the script and language at hand allows selecting the correct language model to employ~\cite{unnikrishnan2009}. While script identification has been widely studied in document analysis, it remains an almost unexplored problem for scene text. In contrast to document images, scene text presents a set of specific challenges, stemming from the high variability in terms of perspective distortion, physical appearance, variable illumination and typeface design. At the same time, scene text comprises typically a few words, contrary to longer text passages available in document images.

Current end-to-end systems for scene text reading~\cite{bissacco2013,Jaderberg2014} assume single script and language inputs given beforehand, i.e. provided by the user, or inferred from available meta-data.  
The unconstrained text understanding problem for large collections of images from unknown sources has not been considered up to very recently~\cite{Shi2015}.
While there exists some research in script identification of text over complex backgrounds~\cite{gllavata2005script,shivakumara2015}, such methods have been so far limited to video overlaid-text, which presents in general different challenges than scene text.

\begin{figure}[t]
\includegraphics[width=\linewidth]{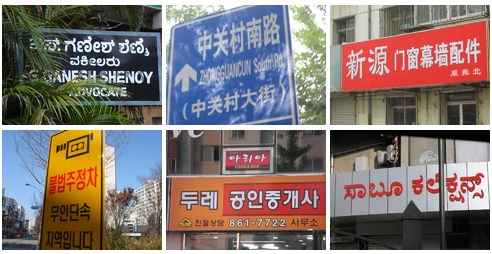}
\caption{Collections of images from unknown sources may contain textual information in different scripts.}
\label{fig:babel_db}
\end{figure}

\begin{figure}[h]
\includegraphics[width=\linewidth]{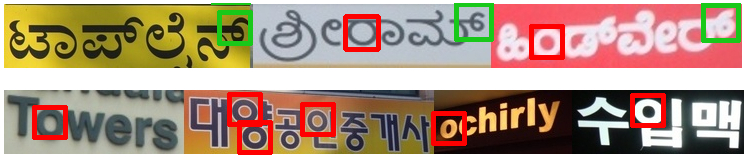}
\caption{Certain stroke-parts (in green) are discriminative for the identification of a particular script (left), while others (in red) can be trivially ignored because are frequent in other classes (right).}
\label{fig:stroke_parts}
\end{figure}

This paper addresses the problem of script identification in natural scene images, paving the road towards true multi-language end-to-end scene text understanding.
Multi-script text exhibits high intra-class variability (words written in the same script vary a lot) and high inter-class similarity (certain scripts resemble each other). Examining text samples from different scripts, it is clear that some stroke-parts are quite discriminative, whereas others can be trivially ignored as they occur in multiple scripts. The ability to distinguish these relevant stroke-parts can be leveraged for recognizing the corresponding script. Figure~\ref{fig:stroke_parts} shows an example of this idea.

The method presented is based on a novel combination of convolutional features~\cite{coates2011analysis} with the Naive-Bayes Nearest Neighbor (NBNN) classifier~\cite{boiman2008defense}. 
 
The key intuition behind the proposed framework is to construct powerful local feature representations and use them within a classifier framework that is able to retain the discriminative power of small image parts.
In this sense, script identification can be seen as a particular case of fine-grained recognition. 
Our work takes inspiration from recent methods in fine-grained recognition that make use of small image patches~\cite{yao2012codebook,krause2014learning}, like NBNN does. Both NBNN and those template-patch based methods implicitly avoid any code-word quantization, in order to avoid loss of discriminability.

Moreover, we propose a novel way to discover the most discriminative per-class stroke-parts (patched) by leveraging the topology of the NBNN search space, providing a weighted image to class metric distance.

The paper also introduces a new benchmark dataset, namely the ``MLe2e'' dataset, for the evaluation of scene text end-to-end reading systems and all intermediate stages such as text detection, script identification and text recognition.
The dataset contains a total of $711$ scene images covering four different scripts (Latin, Chinese, Kannada, and Hangul) and a large variability of scene text samples.

\section{Related Work}
\label{sec:background}
Research in script identification on non traditional paper layouts is scarce, and to the best of our knowledge it has been so far mainly dedicated to video overlaid-text. Gllavatta et al.~\cite{gllavata2005script}, in the first work dealing this task, propose the use of the wavelet transform to detect edges in text line images. Then, they extract a set of low-level edge features, and make use of a K-NN classifier.

Sharma et al.~\cite{sharma2013word} have explored the use of traditional document analysis techniques for video overlaid-text script identification at word level. They analyze three sets of features: Zernike moments, Gabor filters, and a set of hand-crafted gradient features previously used for handwritten character recognition; and propose a number of pre-processing algorithms to overcome the inherent challenges of video. In their experiments the combination of super resolution, gradient features, and a SVM classifier perform significantly better that the other combinations.

Shivakumara et al.~\cite{shivakumara2014gradient,shivakumara2015} rely on skeletonization of the dominant gradients and then analyze the angular curvatures~\cite{shivakumara2014gradient} of skeleton components, and the spatial/structural~\cite{shivakumara2015} distribution of their end, joint, and intersection points to extract a set of hand-crafted features. For classification they build a set of feature templates from train data, and use the Nearest Neighbor rule for classifying scripts at word~\cite{shivakumara2014gradient} or text block~\cite{shivakumara2015} level.

As said before, all these methods have been designed specifically for video overlaid-text, which presents in general different challenges than scene text. Concretely, they mainly rely in accurate edge detection of text components and this is not always feasible in scene text. 

A much more recent approach to scene text script identification is provided by Shi et al.~\cite{Shi2015} where the authors propose the Multi-stage Spatially-sensitive Pooling Network (MSPN). The MSPN network overcomes the limitation of having a fixed size input in traditional Convolutional Neural Networks by pooling along each row of the intermediate layers' outputs by taking the maximum (or average) value in each row.

Our work takes inspiration from recent methods in fine-grained recognition. In particular, Krause et al.~\cite{krause2014learning} focus on learning expressive appearance descriptors and localizing discriminative parts. By analyzing images of objects with the same pose they automatically discover which are the most important parts for class discrimination. Yao et al.~\cite{yao2012codebook} obtain image representations by running template matching using a large number of randomly generated image templates. Then they use a bagging-based algorithm to build a classifier by aggregating a set of discriminative yet largely uncorrelated classifiers.

Our method resembles~\cite{yao2012codebook,krause2014learning} in trying to discover the most discriminative parts (or templates) per class. However, in our case we do not assume those discriminative parts to be constrained in space, because the relative arrangement of individual patches in text samples of the same script is largely variable.

\section{Method description}
\label{sec:method}
\begin{figure*}[t]
\includegraphics[width=\linewidth]{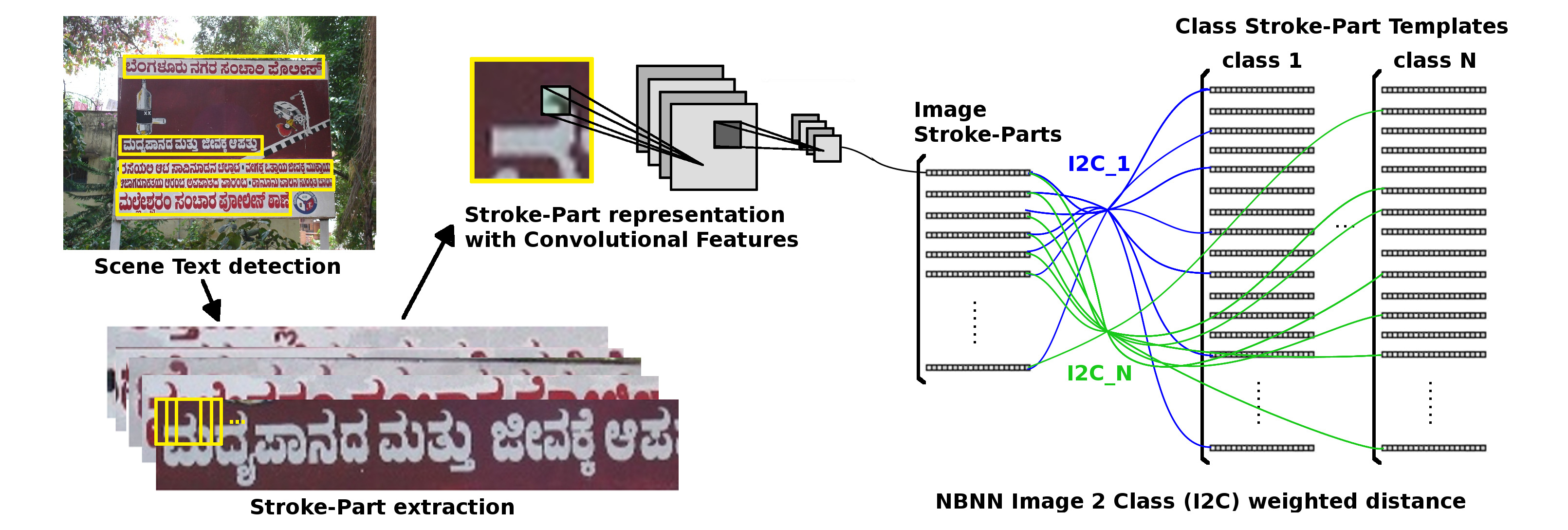}
\caption{Method deploy pipeline: Text lines provided by a text detection algorithm are resized to a fixed height, image patches (stroke-parts) are extracted with a sliding window and fed into a single layer Convolutional Neural Network (CNN). This way, each text line is represented by a variable number of stroke-parts descriptors, that are used to calculate image to class (I2C) distances and classify the input text line using the Naive Bayes Nearest Neighbor (NBNN) classifier.} 
\label{fig:pipeline}
\end{figure*}

Our method for script identification in scene images follows a multi-stage approach.
Given a text line provided by a text detection algorithm, our script identification method proceeds as follows: First we resize the input image to a fixed height of $64$ pixels, but maintaining its original aspect ratio in order to preserve the appearance of stroke-parts.
Second we densely extract $32\times32$ image patches, that we call stroke-parts, with sliding window.
And third, each stroke-part is fed into a single layer Convolutional Neural Network to obtain its feature representation.
These steps are illustrated in Figure~\ref{fig:pipeline} which shows an end-to-end system pipeline incorporating our method (the script-agnostic text detection module is abstracted in a single step as the focus of this paper is on the script identification part).

This way, each input region is represented by a variable number of descriptors (one for each stroke-part), the number of which depends on the length of the input region. Thus, a given text line representation can be seen as a bag of stroke-part descriptors. However, in our method we do not make use of the Bag of visual Words model, as the quantization process severely degrades informative (rare) descriptors~\cite{boiman2008defense}. Instead we directly classify the text lines using the Naive Bayes Nearest Neighbor classifier.

\subsection{Stroke Part representation with Convolutional Features}

\begin{figure}[h]
\includegraphics[width=\linewidth]{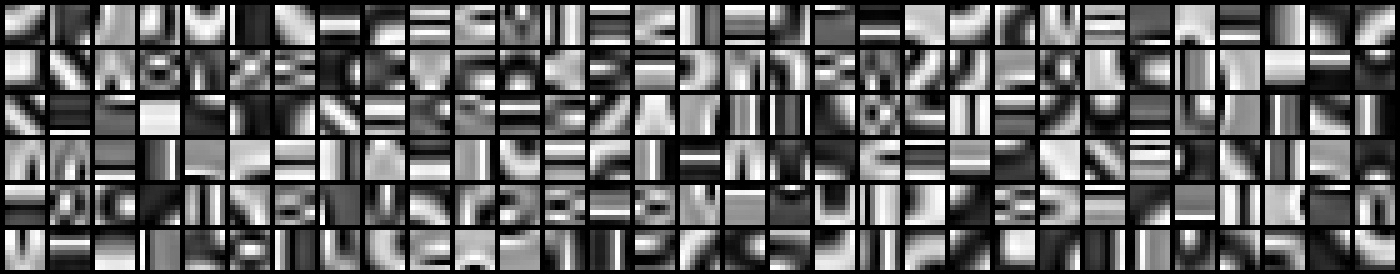}
\caption{Convolution kernels of our single layer network learned with k-means.}
\label{fig:filters}
\end{figure}

Convolutional Features provide the expressive representations of stroke-parts needed in our method. We make use of a single layer Convolutional Neural Network~\cite{coates2011analysis} which provides us with highly discriminative descriptors while not requiring the large amount of training resources typically needed by deeper networks. The weights of the convolutional layer can be efficiently learned using the K-means algorithm.

We adopt a similar design for our network as the one presented in~\cite{Coates2011}. We set the number of convolutional kernels to $256$, the receptive field size to $8\times8$, and we adopt the same non-linear activation function as in~\cite{Coates2011}. After the convolutional layer we stack a spatial average pooling layer to reduce the dimensionality of our representation to $2304$ ($3\times3\times256$). The number of convolutional kernels and kernel sizes of the convolution and pooling layers have been set experimentally, by cross-validation through a number of typical possible values for single-layer networks.

To train the network we first resize all train images to a fixed height, while retaining the original aspect ratio. Then we extract random patches with size equal to the receptive field size, and perform contrast normalization and ZCA whitening~\cite{Kessy2015}. Finally we apply the K-means algorithm to the pre-processed patches in order to learn the $K=256$ convolutional kernels of the CNN. Figure~\ref{fig:filters} depicts a subset of the learned convolutional kernels where it can be appreciated their resemblance to small elementary stroke-parts.

Once the network is trained, the convolutional feature representation of a stroke-part is obtained by feeding its $32\times32$ pixels image patch into the CNN input, after contrast normalization and ZCA whitening.

A key difference of our work with~\cite{Coates2011}, and in general with the typical use of CNN feature representations, is that we do not aim at representing the whole input image with a single feature vector, but instead we extract a set of convolutional features from small parts in a dense fashion. The number of features per image vary according to its aspect ratio. Notice that the typical use of a CNN, resizing the input images to a fixed aspect ratio, is not appealing in our case because it may produce a significant distortion of the discriminative parts of the image that are characteristic of its class.

\subsection{Naive-Bayes Nearest Neighbor}

The Naive-Bayes Nearest Neighbor (NBNN) classifier~\cite{boiman2008defense} is a natural choice in our pipeline because it computes direct Image-to-Class (I2C) distances without any intermediate descriptor quantization. Thus, there is no loss in the discriminative power of the stroke-part representations.
Moreover, having classes with large diversity encourages the use of I2C distances instead of measuring Image-to-Image similarities.

All stroke-parts extracted from the training set images provide the templates that populate the NBNN search space.

In NBNN the I2C distance $d_{I2C}(I,C)$ is computed as $\sum_{i=1}^n{\|d_i~-~NN_C(d_i)\|}^2$, where $d_i$ is the i-th descriptor of the query image $I$, and $NN_C(d_i)$ is the Nearest Neighbor of $d_i$ in class $C$. Then the NBNN classifies the query image to the class $\hat{C}$ with lower I2C distance, i.e. $\hat{C}~=~argmin_C~~d_{I2C}(I,C)$. Figure~\ref{fig:pipeline} shows how computation of I2C distances in our pipeline reduces to $N \times n$ Nearest Neighbor searches, where $N$ is the number of classes and $n$ is the number of descriptors in the query image. To efficiently search for the $NN_C(d_i)$ we make use of the Fast Approximate Nearest Neighbor kd-tree algorithm described in~\cite{muja2009fast}.

\subsection{Weighting per class stroke-part templates by their importance}
\label{sec:weighting}

When measuring the I2C distance $d_{I2C}(I,C)$ it is possible to use a weighted distance function which weights each stroke-part template in the train dataset accounting for its discriminative power. The weighted I2C is then computed as \\* $\sum_{i=1}^n{(1-w_{NN_C(d_i)})\|d_i~-~NN_C(d_i)\|}^2$, where $w_{NN_C(d_i)}$ is the weight of the Nearest Neighbor of $d_i$ of class $C$. 
The weight assigned to each template reflects the ability to discriminate against the class that the template can discriminate best.

We learn the weights associated to each stroke-part template as follows. First, for each template we search for the maximum distance to any of its Nearest Neighbors in all classes except their own class, then we normalize these values in the range $[0,1]$ dividing by the largest distance encountered over all templates. This way, templates that are important in discriminating one class against, at least, one other class have lower contribution to the I2C distance when they are matched as $NN_C$ of one of the query image's parts.

\section{Experiments}
\label{sec:experiments}
All reported experiments were conducted over two datasets, namely the Video Script Identification Competition (CVSI-2015)~\cite{Sharmai2015} dataset and the MLe2e dataset.

The CVSI-2015~\cite{Sharmai2015} dataset comprises pre-segmented video words in ten scripts: English, Hindi, Bengali, Oriya, Gujrathi, Punjabi, Kannada, Tamil, Telegu, and Arabic. The dataset contains about 1000 words for each script and is divided into three parts: a training set ($~60\%$ of the total images), a validation set (10\%), and a test set (30\%). Text is extracted from various video sources (news, sports etc.) and, while  it contains a few instances of scene text, it covers mainly overlay video text.

\subsection{The MLe2e dataset}

This paper introduces the MLe2e multi-script dataset for the evaluation of scene text end-to-end reading systems and all intermediate stages: text detection, script identification and text recognition. The MLe2e dataset has been harvested from various existing scene text datasets for which the images and ground-truth have been revised in order to make them homogeneous. The original images come from the following datasets: Multilanguage(ML)~\cite{Pan2009} and MSRA-TD500~\cite{yao2012detecting} contribute Latin and Chinese text samples, Chars74K~\cite{deCampos09} and MSRRC~\cite{kumar2013multi} contribute Latin and Kannada samples, and KAIST~\cite{lee2010} contributes Latin and Hangul samples.
MLe2e is available at \url{http://158.109.8.43/script_identification/}.

In order to provide a homogeneous dataset, all images have been resized proportionally to fit in $640\times480$ pixels, which is the default image size of the KAIST dataset. Moreover, the ground-truth has been revised to ensure a common text line annotation level~\cite{karatzas2014line}. 
During this process human annotators were asked to review all resized images, adding the script class labels to the ground-truth, and checking for annotation consistency: discarding images with unknown scripts or where all text is unreadable (this may happen because images were resized); joining individual word annotations into text line level annotations; discarding images where correct text line segmentation is not clear or cannot be established, and images where a bounding box annotation contains significant parts of more than one script or significant parts of background (this may happen with heavily slanted or curved text). Arabic numerals ($0,..,9$), widely used in combination with many (if not all) scripts, are labeled as follows. A text line containing text and Arabic numerals is labeled as the script of the text it contains, while a text line containing only Arabic numerals is labeled as Latin. 

The MLe2e dataset contains a total of $711$ scene images covering four different scripts (Latin, Chinese, Kannada, and Hangul) and a large variability of scene text samples. The dataset is split into a train and a test set with $450$ and $261$ images respectively. The split was done randomly, but in a way that the test set contains a balanced number of instances of each class (approx. $160$ text lines samples of each script), leaving the rest of the images for the train set (which is not balanced by default). The dataset is suitable for evaluating various typical stages of end-to-end pipelines, such as multi-script text detection, joint detection and script identification, and script identification in pre-segmented text lines. For the latter, the dataset also provides the cropped images with the text lines corresponding to each data split: $1178$ and $643$ images in the train and test set respectively.

\subsection{Script identification in pre-segmented text lines}

First, we study the performance of the proposed method for script identification in pre-segmented text lines. Table~\ref{tab:baseline} show the obtained results with two variants of our method that only differ by the number of pixels used as step size in the sliding window stage ($8$ or $16$). We provide a comparison with three well known image recognition pipelines using Scale Invariant Features~\cite{lowe1999object} (SIFT) in three different encodings: Fisher Vectors, Vector of Locally Aggregated Descriptors (VLAD), and Bag of Words(BoW); and a linear SVM classifier. In all baselines we extract SIFT features at four different scales in sliding window with a step of 8 pixels. For the Fisher vectors we use a 256 visual words GMM, for VLAD a 256 vector quantized visual words, and for BoW 2,048 vector quantized visual words histograms. The step size and number of visual words were set to similar values to our method when possible in order to offer a fair evaluation. These three pipelines have been implemented with the VLFeat~\cite{vedaldi08vlfeat} and liblinear~\cite{liblinear} open source libraries. We also compare against a Local Binary Pattern variant, the SRS-LBP~\cite{nicolaou2015sparse} pooled over the whole image and followed by a simple KNN classifier.

\begin{table}[h]
\begin{center}
\begin{tabular}{|l|c|c|}
\hline
Method & Acc. CVSI & Acc. MLe2e \\
\hline\hline
SIFT + Bag of Words + SVM & 84.38 & 86.45 \\
SIFT + Fisher Vectors + SVM & 94.11 & 88.63 \\
SIFT + VLAD + SVM & 93.92 & 90.19 \\
SRS-LBP + KNN~\cite{nicolaou2015sparse} & 94.20 & 82.71 \\
\hline
Convolutional Features* + NBNN & 95.88 & 84.57 \\
Convolutional Features* + NBNN + weighting & 96.00 & 88.16 \\
Convolutional Features + NBNN & \textbf{97.91} & 89.87 \\
Convolutional Features + NBNN + weighting & 96.42 & \textbf{91.12} \\
\hline
\end{tabular}
\end{center}
\caption{Script identification accuracy on pre-segmented text lines. Methods marked with an asterisk make use of a 16-pixel step size in the sliding window for stroke-parts extraction.}
\label{tab:baseline}
\end{table}

As shown in Table~\ref{tab:baseline} the proposed method outperforms all baseline methods. 

The contribution of weighting per class Stroke Part Templates by their importance as explained in section~\ref{sec:weighting} is significant in the MLe2e dataset, especially when using larger steps for the sliding window stroke-parts extraction, while producing a small accuracy discount in the CVSI-2015 dataset. 
Our interpretation of these results relates to the distinct nature of the two datasets. On the one hand, CVSI's overlaid-text variability is rather limited compared with that found in the scene text of MLe2e, and secondly in CVSI the number of templates is much larger (about one order of magnitude). Thus, our weighting strategy is more indicated for the MLe2e case where important (discriminative) templates may fall isolated in some region of the NBNN search space.

Table~\ref{tab:cvsi} shows the overall performance comparison of our method with the participants in the ICDAR2015 Competition on Video Script Identification (CVSI 2015)~\cite{Sharmai2015}. Methods labeled as CVC-1 and CVC-2 correspond to the method described in this paper, however notice that as participants in the competition we used the configuration with a 16 pixels step sliding window, i.e. CVC-1 and CVC-2 correspond respectively to "Convolutional Features* + NBNN" and "Convolutional Features* + NBNN + weighting" in Table~\ref{tab:baseline}. As can be appreciated in the table, adding the configuration with an 8 pixels step our method ranks second in the table, only $1\%$ under the winner of the competition.

\begin{table}[h]
\begin{center}
\begin{tabular}{|l|c|c|}
\hline
Method & CVSI (Overall performance) \\
\hline\hline
Google & \textbf{98.91} \\
\textbf{Ours (8 pixel step)} & 97.91 \\
HUST~\cite{Shi2015} & 96.69 \\
\textbf{CVC-2} & 96.00 \\
\textbf{CVC-1} & 95.88 \\
C-DAC & 84.66 \\
CUK & 74.06 \\
\hline
\end{tabular}
\end{center}
\caption{Overall classification performance comparison with participants in the ICDAR2015 competition on video script identification CVSI considering all the ten scripts~\cite{Sharmai2015}.}
\label{tab:cvsi}
\end{table}

The CVSI-2015 competition winner (Google) makes use of a deep convolutional network for class prediction that is trained using data-augmentation techniques. Our method demonstrates competitive performance with a shallower design that implies a much faster and attainable training procedure.

\begin{figure}[t]
\includegraphics[width=\linewidth]{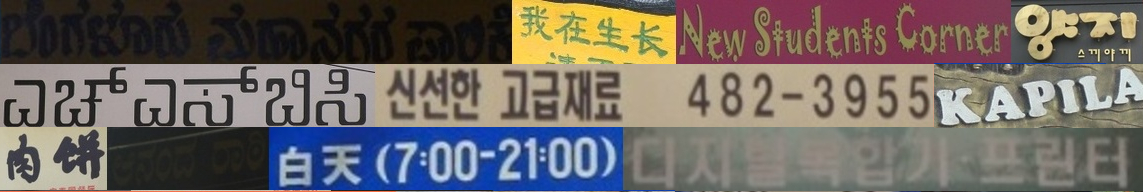}
\caption{A selection of 
misclassified samples by our method: low contrast images, rare font types, degraded text, letters mixed with numerals, etc.}
\label{fig:babel_errors}
\end{figure}

\subsection{Joint text detection and script identification in scene images}

In this experiment we evaluate the performance of a complete pipeline for detection and script identification in its joint ability to detect text lines in natural scene images and properly recognizing their scripts. The key interest of this experiment is to study the performance of the proposed script identification algorithm when realistic, non-perfect, text localization is available.

Most text detection pipelines are trained explicitly for a specific script (typically English) and generalize pretty badly to the multi-script scenario. We have chosen to use here the script-agnostic method of Gomez et al.~\cite{Gomez2014}, which is designed for multi-script text detection that generalizes well to any script.
The method detects character candidates using the Maximally Stable Extremal Regions (MSER)~\cite{Matas2004}, and builds different hierarchies where the initial regions are grouped by agglomerative clustering, using complementary similarity measures. In such hierarchies each node defines a possible text hypothesis. Then, an efficient classifier, using incrementally computable descriptors, is used to walk each hierarchy and select the nodes with larger text-likelihood.

In this paper script identification is performed at the text line level, because segmentation into words is largely script-dependent. Notice however that in some cases, by the intrinsic nature of scene text, a text line provided by the text detection module may correspond to a single word, so we must deal with a large variability on the length of provided text lines. 
The experiments are performed over the new MLe2e dataset.

For evaluation of the joint text detection and script identification task in the MLe2e dataset we propose the use of simple two-stage framework. First, localization is assessed based on the Intersection-over-Union (IoU) metric between detected and ground-truth regions, as commonly used in object recognition tasks~\cite{everingham2014pascal} and the recent ICDAR 2015 competition\footnote{\url{http://rrc.cvc.uab.es}}. Second, the predicted script is verified against the ground-truth. A detected bounding box is thus considered a True Positive if has a IoU $>0.5$ with a bounding box in the ground-truth and the predicted script is correct.

The localization-only performance, corresponding to the first stage of the evaluation, yields an F-score of $0.58$ (Precision of $0.54$ and Recall of $0.62$). This defines the upper-bound for the joint task.

The two stage evaluation, including script identification using the proposed method, achieves an F-score of $0.51$, analyzed into a Precision of $0.48$ and a Recall of $0.55$. 

The results demonstrate that the proposed method for script identification is effective even when the text region is badly localized, as long as part of the text area is within the localized region. This extends to regions that did not pass the $0.5$ IoU threshold, but had their script correctly identified. Such a behavior is to be expected, due to the way our method treats local information to decide on a script class. This opens the possibility to make use of script identification to inform and / or improve the text localization process. The information of the identified script can be used to refine the detections. 

\subsection{Cross-domain performance and confusion in Latin-only datasets}

In this experiment we evaluate the cross-domain performance of learned stroke-part templates from one dataset to the other. We evaluate on the CVSI test set using the templates learned in the MLe2e train set, and the other way around, by measuring classification accuracy only for the two common script classes: Latin and Kannada. Finally, we evaluate the misclassification error of our method using script-part templates learned from both datasets over a third Latin-only dataset. For this experiment we use the ICDAR2013 scene text dataset~\cite{karatzas2013icdar} which provides cropped word images of English text, and measure the classification accuracy of our method. Table~\ref{tab:cross-domain} shows the results of these experiments.

\begin{table}[h]
\begin{center}
\begin{tabular}{|l|c|c||r|}
\hline
Method & CVSI & MLe2e & ICDAR\\
\hline\hline
Conv. Feat. + NBNN + W (CVSI) & 95.11 & 45.98 & 43.40\\
\hline\hline
Conv. Feat. + NBNN + W (MLe2e) & 70.22 & 91.67 & 94.70\\
\hline
\end{tabular}
\end{center}
\caption{Cross-domain performance of our method measured by training/testing in different datasets.}
\label{tab:cross-domain}
\end{table}

From the above table we can see how features learned on the MLe2e dataset are much better in generalizing to other datasets.  
In fact, this is an expected result, because the domain of overlay text in CVSI can be seen as a subdomain of the scene text MLe2e's domain. Since the MLe2e dataset is richer in text variability, e.g. in terms of perspective distortion, physical appearance, variable illumination and typeface designs, makes script identification on this dataset a more difficult problem, but also more indicated if one wants to learn effective cross-domain stroke-part descriptors. 
Significantly important is the result obtained in the English-only ICDAR dataset which is near $95\%$. 
This demonstrates that our method is able to learn discriminative stroke-part representations that are not dataset-specific. 
It is important to notice that the rows in Table~\ref{tab:cross-domain} are not directly comparable as both models have different numbers of classes, 10 in the case of training over the CVSI dataset and 4 in the case of the MLe2e.
However, the experiment is relevant when comparing the performance of a learned model on datasets different from the one used for training. In this sense, the obtained results show a clear weakness of the features learned on the video overlaid text of CVSI for correctly identifying the script in scene text images. On the contrary, features learned in the MLe2e dataset perform very  well in other scene text data (ICDAR), while exhibit an expected but acceptable decrement in performance in video overlaid text (CVSI-2015).

\begin{figure}[t]
\includegraphics[width=\linewidth]{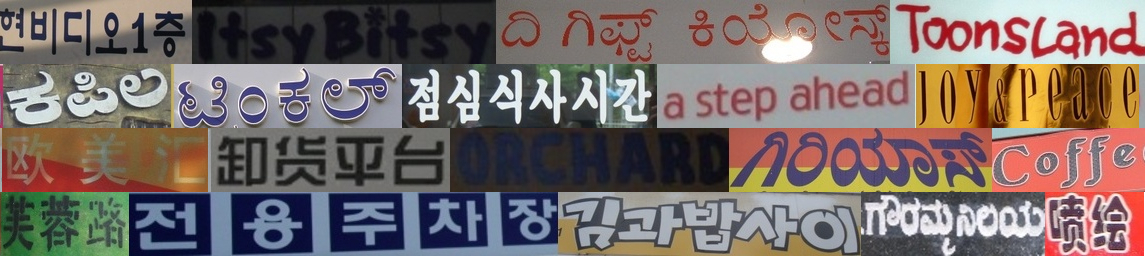}
\caption{Examples of correctly classified instances}
\label{fig:babel_ok}
\end{figure}


\section{Conclusion}
\label{sec:conclusion}
A novel method for script identification in natural scene images was presented. The method combines the expressive representation of convolutional features and the fine-grained classification characteristics of the Naive-Bayes Nearest Neighbor classifier. 
In addition, a new public benchmark dataset for the evaluation of all stages of end-to-end scene text reading systems was introduced. 
Experiments done in this new dataset and the CVSI video overlay dataset exhibit state of the art accuracy rates in comparison to a number methods, including the participants in the CVSI-2015 competition and standard image recognition pipelines. 
Our work demonstrates the viability of script identification in natural scene images, paving the road towards true multi-language end-to-end scene text understanding.

Source code of our method and the MLe2e dataset are available online at \url{http://158.109.8.43/script_identification/}.

\section{Acknowledgments}
This work was supported by the Spanish project TIN2014-52072-P, the fellowship RYC-2009-05031, and the Catalan govt scholarship 2014FI\_B1-0017.
\small
\bibliographystyle{IEEEtran}
\bibliography{IEEEabrv,GomezKaratzas_das16}

\begin{thebibliography}{10}
\providecommand{\url}[1]{#1}
\csname url@samestyle\endcsname
\providecommand{\newblock}{\relax}
\providecommand{\bibinfo}[2]{#2}
\providecommand{\BIBentrySTDinterwordspacing}{\spaceskip=0pt\relax}
\providecommand{\BIBentryALTinterwordstretchfactor}{4}
\providecommand{\BIBentryALTinterwordspacing}{\spaceskip=\fontdimen2\font plus
\BIBentryALTinterwordstretchfactor\fontdimen3\font minus
  \fontdimen4\font\relax}
\providecommand{\BIBforeignlanguage}[2]{{%
\expandafter\ifx\csname l@#1\endcsname\relax
\typeout{** WARNING: IEEEtran.bst: No hyphenation pattern has been}%
\typeout{** loaded for the language `#1'. Using the pattern for}%
\typeout{** the default language instead.}%
\else
\language=\csname l@#1\endcsname
\fi
#2}}
\providecommand{\BIBdecl}{\relax}
\BIBdecl

\bibitem{unnikrishnan2009}
R.~Unnikrishnan and R.~Smith, ``Combined script and page orientation estimation
  using the tesseract ocr engine,'' in \emph{MOCR}, 2009.

\bibitem{bissacco2013}
A.~Bissacco, M.~Cummins, Y.~Netzer, and H.~Neven, ``Photoocr: Reading text in
  uncontrolled conditions,'' in \emph{ICCV}, 2013.

\bibitem{Jaderberg2014}
M.~Jaderberg, A.~Vedaldi, and A.~Zisserman, ``Deep features for text
  spotting,'' in \emph{ECCV}, 2014.

\bibitem{Shi2015}
B.~Shi, C.~Yao, C.~Zhang, X.~Guo, F.~Huang, and X.~Bai, ``Automatic script
  identification in the wild,'' \emph{ICDAR}, 2015.

\bibitem{gllavata2005script}
J.~Gllavata and B.~Freisleben, ``Script recognition in images with complex
  backgrounds,'' in \emph{SPIT}, 2005.

\bibitem{shivakumara2015}
P.~Shivakumara, Z.~Yuan, D.~Zhao, T.~Lu, and C.~L. Tan, ``New
  gradient-spatial-structural features for video script identification,''
  \emph{CVIU}, 2015.

\bibitem{coates2011analysis}
A.~Coates, A.~Y. Ng, and H.~Lee, ``An analysis of single-layer networks in
  unsupervised feature learning,'' in \emph{AIStats}, 2011.

\bibitem{boiman2008defense}
O.~Boiman, E.~Shechtman, and M.~Irani, ``In defense of nearest-neighbor based
  image classification,'' in \emph{CVPR}, 2008.

\bibitem{yao2012codebook}
B.~Yao, G.~Bradski, and L.~Fei-Fei, ``A codebook-free and annotation-free
  approach for fine-grained image categorization,'' in \emph{CVPR}, 2012.

\bibitem{krause2014learning}
J.~Krause, T.~Gebru, J.~Deng, L.-J. Li, and L.~Fei-Fei, ``Learning features and
  parts for fine-grained recognition,'' in \emph{ICPR}, 2014.

\bibitem{sharma2013word}
N.~Sharma, S.~Chanda, U.~Pal, and M.~Blumenstein, ``Word-wise script
  identification from video frames,'' in \emph{ICDAR}, 2013.

\bibitem{shivakumara2014gradient}
P.~Shivakumara, N.~Sharma, U.~Pal, M.~Blumenstein, and C.~L. Tan,
  ``Gradient-angular-features for word-wise video script identification,'' in
  \emph{ICPR}, 2014.

\bibitem{Coates2011}
A.~Coates, B.~Carpenter, C.~Case, S.~Satheesh, B.~Suresh, T.~Wang, D.~Wu, and
  A.~Ng, ``Text detection and character recognition in scene images with
  unsupervised feature learning,'' in \emph{Proc. ICDAR}, 2011.

\bibitem{Kessy2015}
A.~{Kessy}, A.~{Lewin}, and K.~{Strimmer}, ``{Optimal whitening and
  decorrelation},'' \emph{arXiv preprint arXiv:1512.00809}, 2015.

\bibitem{muja2009fast}
M.~Muja and D.~G. Lowe, ``Fast approximate nearest neighbors with automatic
  algorithm configuration.'' \emph{VISAPP}, 2009.

\bibitem{Sharmai2015}
N.~Sharmai, R.~Mandal, M.~Blumenstein, and U.~Pal, ``{ICDAR} 2015 competition
  on video script identification ({CVSI}-2015),'' \emph{ICDAR}, 2015.

\bibitem{Pan2009}
Y.-F. Pan, X.~Hou, and C.-L. Liu, ``Text localization in natural scene images
  based on conditional random field,'' in \emph{Proc. ICDAR}, 2009.

\bibitem{yao2012detecting}
C.~Yao, X.~Bai, W.~Liu, Y.~Ma, and Z.~Tu, ``Detecting texts of arbitrary
  orientations in natural images,'' in \emph{CVPR}, 2012.

\bibitem{deCampos09}
T.~E. de~Campos, B.~R. Babu, and M.~Varma, ``Character recognition in natural
  images,'' in \emph{ICCVTA}, 2009.

\bibitem{kumar2013multi}
D.~Kumar, M.~Prasad, and A.~Ramakrishnan, ``Multi-script robust reading
  competition in icdar 2013,'' in \emph{MOCR}, 2013.

\bibitem{lee2010}
S.~Lee, M.~S. Cho, K.~Jung, and J.~H. Kim, ``Scene text extraction with edge
  constraint and text collinearity.'' in \emph{ICPR}, 2010.

\bibitem{karatzas2014line}
D.~Karatzas, S.~Robles, and L.~Gomez, ``An on-line platform for ground truthing
  and performance evaluation of text extraction systems,'' in \emph{DAS}, 2014.

\bibitem{lowe1999object}
D.~G. Lowe, ``Object recognition from local scale-invariant features,'' in
  \emph{ICCV}, 1999.

\bibitem{vedaldi08vlfeat}
A.~Vedaldi and B.~Fulkerson, ``{VLFeat}: An open and portable library of
  computer vision algorithms,'' \url{http://www.vlfeat.org/}, 2008.

\bibitem{liblinear}
R.-E. Fan, K.-W. Chang, C.-J. Hsieh, X.-R. Wang, and C.-J. Lin, ``{LIBLINEAR}:
  A library for large linear classification,'' \emph{JMLR}, 2008.

\bibitem{nicolaou2015sparse}
A.~Nicolaou, A.~D. Bagdanov, M.~Liwicki, and D.~Karatzas, ``Sparse radial
  sampling lbp for writer identification,'' \emph{ICDAR}, 2015.

\bibitem{Gomez2014}
L.~Gomez and D.~Karatzas, ``A fast hierarchical method for multi-script and
  arbitrary oriented scene text extraction,'' \emph{arXiv preprint
  arXiv:1407.7504}, 2014.

\bibitem{Matas2004}
J.~Matas, O.~Chum, M.~Urban, and T.~Pajdla, ``Robust wide-baseline stereo from
  maximally stable extremal regions,'' \emph{Image and Vision Computing}, 2004.

\bibitem{everingham2014pascal}
M.~Everingham, S.~A. Eslami, L.~Van~Gool, C.~K. Williams, J.~Winn, and
  A.~Zisserman, ``The pascal visual object classes challenge: A
  retrospective,'' \emph{IJCV}, 2014.

\bibitem{karatzas2013icdar}
D.~Karatzas, F.~Shafait, S.~Uchida, M.~Iwamura, S.~R. Mestre, J.~Mas, D.~F.
  Mota, J.~A. Almazan, L.~P. de~las Heras \emph{et~al.}, ``Icdar 2013 robust
  reading competition,'' in \emph{ICDAR}, 2013.

\end{thebibliography}

\end{document}